\documentclass{llncs}
\usepackage{graphicx}
\usepackage{amssymb}
\usepackage{amsmath}
\usepackage{bbm}
\usepackage{caption} 

\begin{document}

\title{Long-Term Outlier Prediction \\ Through Outlier Score Modeling}

\author{
Yuma~Aoki\inst{1} \and
Joon~Park\inst{2} \and
Koh~Takeuchi\inst{2} \and
Hisashi~Kashima\inst{2} \and \\
Shinya~Akimoto\inst{3} \and
Ryuichi~Hashimoto\inst{3} \and
Takahiro~Adachi\inst{3} \and
Takeshi~Kishikawa\inst{3} \and
Takamitsu~Sasaki\inst{3} 
}
\authorrunning{Y. Aoki et al.}
\institute{Nara Institute of Science and Technology \and
Kyoto University \and
Panasonic Holdings Corporation}

\maketitle

\begin{abstract}
This study addresses an important gap in time series outlier detection by proposing a novel problem setting: long-term outlier prediction. Conventional methods primarily focus on immediate detection by identifying deviations from normal patterns. As a result, their applicability is limited when forecasting outlier events far into the future. To overcome this limitation, we propose a simple and unsupervised two-layer method that is independent of specific models. The first layer performs standard outlier detection, and the second layer predicts future outlier scores based on the temporal structure of previously observed outliers. This framework enables not only pointwise detection but also long-term forecasting of outlier likelihoods. Experiments on synthetic datasets show that the proposed method performs well in both detection and prediction tasks. These findings suggest that the method can serve as a strong baseline for future work in outlier detection and forecasting.
\end{abstract}

\section{Introduction}

Outlier detection in time series data has been widely studied due to its relevance in many real-world applications, including production monitoring and network security~\cite{intro1,blazquez2021review}.
Outliers are rare and irregular events that are not frequently observed, which makes them well suited to unsupervised learning settings.
Most existing methods for time series outlier detection attempt to identify normal patterns and detect anomalies as deviations from these patterns~\cite{intro2}.

Despite their success in identifying immediate anomalies, these conventional methods typically focus on short-term deviations and do not support the prediction of outlier occurrences in the distant future.
This limitation significantly reduces their usefulness in situations where long-term forecasting and interpretation are important.

This paper focuses on the unsupervised prediction of outliers over long time horizons, which addresses a critical gap in the existing literature.
Predicting outliers far in advance is fundamentally difficult, because most current methods detect anomalies based on deviations of present observations from recent norms.
To the best of our knowledge, there are almost no existing studies that have explored the long-term prediction of outliers.

To solve this problem, it is necessary to go beyond isolated outlier detection and capture temporal dependencies among outlier events.
Such dependencies arise in many real-world scenarios.
For instance, seasonal patterns in infectious diseases, periodic failures in industrial equipment, and fluctuations in network traffic often follow predictable cycles.
In multivariate time series, delayed effects across variables may also occur.
In the financial domain, for example, a sharp change in the stock price of one company can influence the prices of related companies after a certain delay.
Similarly, in manufacturing, a failure in an upstream process may lead to irregularities in downstream operations.

\begin{figure}[t]
    \centering
    \includegraphics[keepaspectratio, scale=0.35]{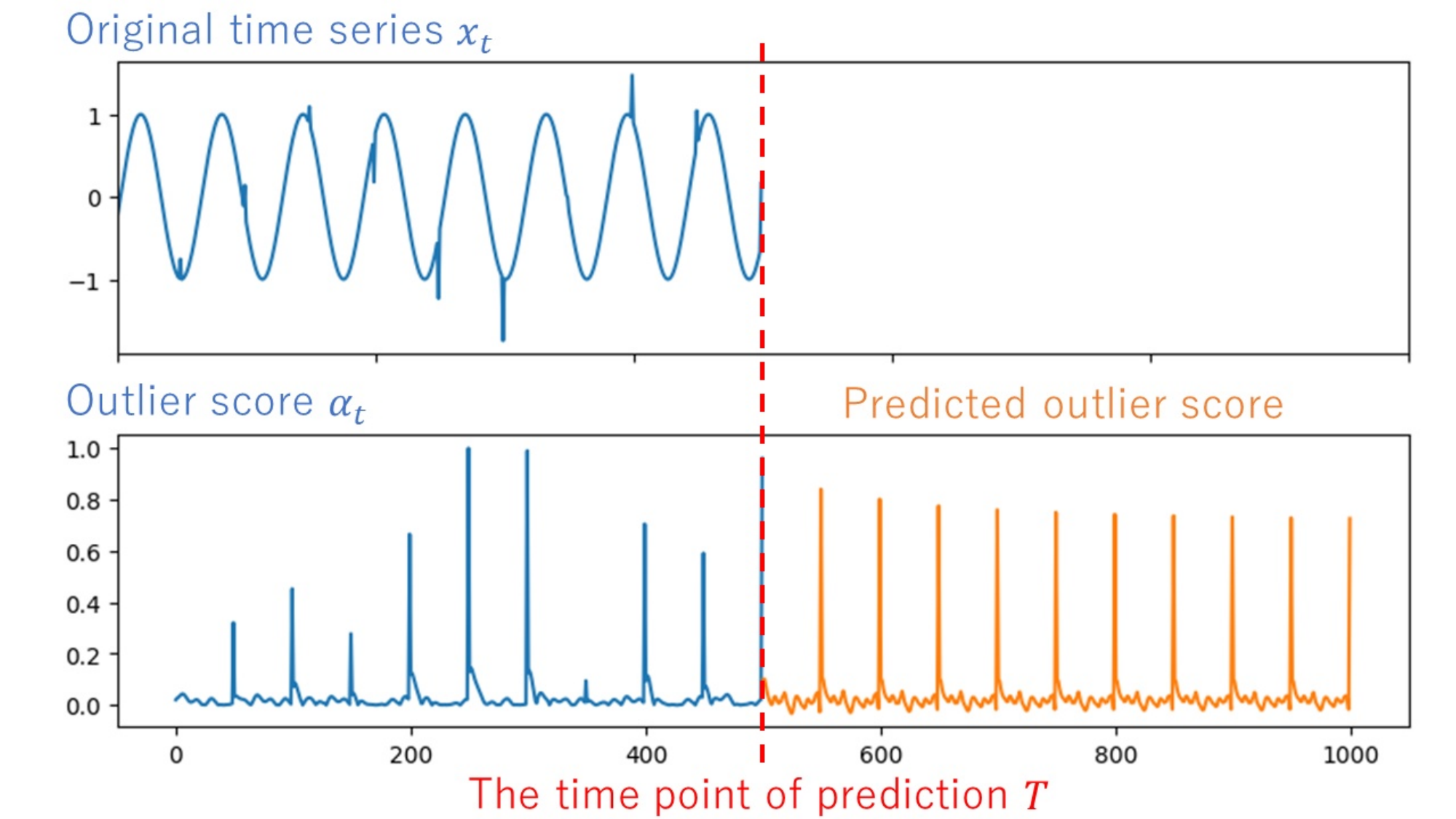}
    \caption{Illustrative example of the proposed approach. The time series (including outliers) is given up to time $T = 500$. Conventional methods calculate outlier scores using only past values, which prevents the detection of future anomalies. However, if a regular pattern such as periodicity exists in the outlier scores, this structure can be used to anticipate future scores.}
    \label{fig:example}
\end{figure}

Our basic idea is illustrated in Fig.~\ref{fig:example}.
Consider a time series generated from the periodic function $x_t = \sin (t/10)$, where random noise (representing outliers) is injected every 50 time steps from the normal distribution $\mathcal{N}(0,1)$.
At time $T$, the values up to that point are known (Fig.~\ref{fig:example}, top).
Conventional outlier detection assigns scores to these past values by comparing them with their expected values (bottom-left, blue line).
If one observes that the resulting outlier scores increase periodically, it becomes possible to anticipate similar increases in the future (bottom-right, orange line).

We propose a simple two-layer framework that is agnostic to the choice of model.
The first layer performs conventional outlier detection and outputs an outlier score at each time point.
The second layer takes the sequence of outlier scores and learns their temporal structure in order to forecast future scores.
This structure enables long-term outlier prediction based on past detection results.
Each layer can be implemented using any standard time series model.

To evaluate the effectiveness of the proposed approach, we conduct experiments using both synthetic and real datasets with regular outlier patterns.
We examine whether the method can correctly anticipate the occurrence of artificially injected anomalies that are temporally or cross-dimensionally related.
The experimental results suggest that the method works well in both univariate and multivariate settings, and serves as a strong baseline for future research.

The contributions of this paper are as follows:
\begin{enumerate}
    \item We define a novel task, long-term outlier prediction, which aims to forecast anomalies beyond the immediate or near-future horizon.
    \item We propose a simple two-layer solution. The first layer detects anomalies at each time point, and the second layer forecasts future outlier scores based on learned temporal dependencies.
    \item We design an experimental framework and show that the proposed method performs well across various settings and evaluation metrics.
\end{enumerate}

\section{Related Work}

This section reviews research related to outlier detection and time-series modeling, which are the foundations of this study. We also discuss their intersection, namely, outlier detection for time-series data. These topics have a long history, extending beyond the recent developments in deep learning. Therefore, we focus on key ideas relevant to the current study rather than providing a comprehensive survey.

\subsection{Outlier Detection}

Outlier detection~\cite{hodge2004survey,boukerche2020outlier} is a long-standing and fundamental problem in machine learning and data mining. It has attracted considerable attention from both academic and industrial communities, due to its theoretical significance and practical importance in various domains.

One of the main challenges in this area is the scarcity of labeled anomaly data. Because anomalous events are rare and often unknown in advance, supervised learning is typically infeasible. As a result, most methods adopt unsupervised learning approaches.

Unsupervised methods generally detect outliers by measuring deviations from the patterns of normal data. Classic examples include the local outlier factor (LOF)~\cite{breunig2000lof}, which assesses abnormality based on local density, and methods based on the likelihood of observed data under a probabilistic model, such as a Gaussian mixture model~\cite{yamanishi2002unifying}. 

In recent years, deep learning-based approaches have emerged as a dominant trend~\cite{chalapathy2019deep}. Among them, autoencoder-based methods have gained popularity by using reconstruction error as an outlier score~\cite{sakurada2014anomaly}.

Given the large number of existing methods, we refer the reader to comprehensive surveys for a broader overview~\cite{chandola2009anomaly,chalapathy2019deep}.

\subsection{Time-series Modeling}

Time-series modeling has a long tradition in statistics and has continued to evolve through advancements in machine learning and data mining.

Early methods include parametric models such as the ARMA family and state-space models~\cite{hamilton2020time,shumway2000time}. With the development of deep learning, neural network-based models have become increasingly popular~\cite{lim2021time}.

Classical Recurrent Neural Networks (RNNs) have been extended to more powerful architectures such as Long Short-Term Memory (LSTM)~\cite{hochreiter1997long} and Gated Recurrent Units (GRU)~\cite{cho2014learning}, which are capable of modeling longer temporal dependencies. Recently, Transformer architectures~\cite{wen2022transformers}, which were originally introduced for natural language processing, have also been adopted in time-series tasks with promising results.

\subsection{Time-series Outlier Detection}

Outlier detection is often required in real-time monitoring systems, such as those used in manufacturing or network operations, where the data naturally comes in the form of time series from sensors and logs.

There are two major approaches to detecting outliers in time-series data~\cite{intro1,blazquez2021review}. One approach divides the series into smaller segments and detects outliers within each fragment. The other approach builds predictive models using time-series techniques and flags observations as outliers when they significantly deviate from the model's prediction~\cite{intro1}. In this work, we follow the latter approach.

\vspace{5mm}
As mentioned above, substantial research has been conducted on both outlier detection and time-series modeling. Their integration has also been actively explored. However, to the best of our knowledge, there has been no prior work that focuses on predicting the occurrence of outliers far into the future. This study presents a simple and novel framework to address this largely unexplored problem.

\section{Proposed Method}

We begin by defining the problem of long-term outlier prediction, which is the main focus of this study. We then present a simple two-layer framework to address this problem. The proposed method combines a standard outlier detector with a time-series forecasting model.

\subsection{Long-term Outlier Prediction Problem}

Let the time series data be $x_1, x_2, \ldots, x_t, \ldots$, where $x_t$ ($t = 1, 2, \ldots$) is a $D$-dimensional real-valued observation at time $t$. When $D = 1$, the series is univariate; otherwise, it is multivariate.

We are interested in detecting irregular or anomalous behavior in this time series. To quantify the degree of abnormality at each time point, we define an outlier score $\alpha_t$, which is also a $D$-dimensional vector. The $j$-th element of $\alpha_t$ reflects the degree of anomaly in the $j$-th dimension of $x_t$.

The goal is to predict the outlier score $\alpha_{T+\tau}$ at a future time $T+\tau$, given only the observations $x_1, x_2, \ldots, x_T$ up to the current time $T$.

In conventional unsupervised outlier detection, the outlier score $\alpha_T$ is computed using only the data available up to time $T$. For example, it is often defined as the deviation between the actual observation $x_T$ and a prediction $\hat{x}_T$ based on past values.

In contrast, our problem requires predicting $\alpha_{T+\tau}$ at a future time. This must be done without access to any data beyond time $T$. When $\tau = 0$, our setting reduces to the standard outlier detection problem.

One might think that predicting $x_{T+\tau}$ would be sufficient, but this is not the case. Determining whether $x_{T+\tau}$ is an outlier requires comparing it with the actual value observed at that time. However, since $x_{T+\tau}$ is in the future, this value is unknown at the current time $T$.

Because of this limitation, some assumption must be made in order to forecast future outliers. In this study, we assume that the timing of outlier events follows certain patterns. While the specific values of future anomalies may not be predictable, we hypothesize that the timing of their occurrence depends on the timing of past outliers.

\subsection{A Simple Two-layer Approach}

Outlier detection typically identifies events that deviate from expected patterns. Predicting such events in advance requires strong assumptions. In particular, we assume that outlier events occur in a temporally structured manner. For example, an outlier may be triggered after a fixed delay due to a hidden periodic pattern. Alternatively, an outlier may be followed by another outlier after a certain time interval.

Such patterns are not necessarily confined to a single variable. In multivariate time series, one variable may influence another, causing delayed cross-dimensional anomalies.

Based on this observation, we propose a two-layer framework for long-term outlier prediction. The framework consists of an outlier detection layer and an outlier score prediction layer. The first layer identifies anomalies at each time point. The second layer learns temporal patterns among these anomalies and forecasts future outlier scores.

\subsubsection*{Outlier detection layer:}

Given the observations $x_1, x_2, \ldots, x_T$ up to time $T$, we apply a standard outlier detection model to compute the corresponding outlier scores $\alpha_1, \alpha_2, \ldots, \alpha_T$. This model can be either pointwise (using only $x_t$) or contextual (using past observations), depending on the application.

The choice of outlier detection method is flexible. Any algorithm that outputs a meaningful anomaly score for each observation can be used. The specific definition of an outlier depends on the application and is not restricted by our framework.

\subsubsection*{Outlier score prediction layer:}

The second layer takes the sequence of outlier scores $\alpha_1, \alpha_2, \ldots, \alpha_T$ as input and fits a time-series model $f$ to this sequence. This model is used to predict future outlier scores, including $\alpha_{T+\tau}$.

These predicted scores can be used to trigger alerts or interventions. For instance, if the predicted score exceeds a threshold, the system may report a potential anomaly in advance.

\vspace{5mm}
One might consider applying a time-series model directly to the observed data $x_1, \ldots, x_T$ and using its prediction to identify future anomalies. However, this approach is not feasible in our setting. A time-series model estimates the expected (normal) value at time $T+\tau$, but determining whether this value is anomalous requires comparing it with the actual observation at $T+\tau$, which is not accessible at the time of prediction.

By modeling the sequence of outlier scores directly, we circumvent this issue. Our method enables the forecasting of outlier scores without requiring future observations. This approach allows long-term outlier prediction under the assumption that outlier occurrences exhibit temporal patterns.

\subsection{Training and Prediction}

The training and prediction procedures are described as follows.  
Let $T$ denote the current time, which is the point at which we wish to predict the future occurrence of an outlier.  
For simplicity, we divide the observed time series data $x_1, x_2, \ldots, x_T$ into two segments at time $\lfloor T/2 \rfloor$.  
This split prevents overlap between the data used to train the outlier detection layer and the data used to generate training targets for the outlier score prediction layer.  
Such separation helps avoid potential bias.

\subsubsection*{Training the outlier detection layer:}  
We use the first half of the data, $x_1, x_2, \ldots, x_{\lfloor T/2 \rfloor}$, to fit a time series prediction model $g$.  
If necessary, model selection can be performed by further dividing this segment into training and validation subsets.

Once trained, the model $g$ is applied to the second half of the time series, $x_{\lfloor T/2 \rfloor + 1}, \ldots, x_T$, to compute the outlier scores $\alpha_{\lfloor T/2 \rfloor + 1}, \ldots, \alpha_T$.  
These scores reflect the deviation of the actual observations from the model’s predictions.  
For instance, the outlier score of the $j$-th variable at time $t$ is given by:
\[
    \alpha_t^{(j)} = |x_t^{(j)} - g_t^{(j)}|,
\]
where $g_t^{(j)}$ is the predicted value for the $j$-th dimension at time $t$.

\subsubsection*{Training the outlier score prediction layer:}  
The second layer is trained using the outlier scores obtained from the previous step, that is, the sequence $\alpha_{\lfloor T/2 \rfloor + 1}, \ldots, \alpha_T$.  
We fit a time series model $f$ to this sequence.  
Since outlier scores are fundamentally different from the original observations, the model $f$ does not need to belong to the same class as $g$.

\subsubsection*{Prediction phase:}  
In the prediction phase, the model $f$ is used to forecast future outlier scores beyond time $T$.  
The specific forecasting method depends on the choice of model.  
A straightforward approach is to perform one-step-ahead predictions recursively.  
Each time a new outlier score is predicted, it is fed back into the model as if it were an observed value, and used to generate the next prediction.

\section{Experiments}

To validate the effectiveness of the proposed method, we conduct experiments using both synthetic and real-world time series that contain artificially inserted outliers.

\subsection{Univariate Time Series}

We use two datasets to examine whether the proposed approach can accurately predict the timing of future outliers in univariate time series.

\subsubsection*{Synthetic Dataset:}

The first dataset is synthetically generated from a simple periodic function defined by
\begin{equation}
x_t = \sin(t/10).
\end{equation}
To introduce anomalies, Gaussian noise sampled from $\mathcal{N}(0, 0.5)$ is added every 50 time steps. This results in a time series where outliers occur at regular intervals.

The complete time series consists of 1500 points. The first 500 points are used to train the time series prediction model in the outlier detection layer. After this, the trained model is applied to the next 500 points to compute the corresponding outlier scores. These scores form a new time series, which is then used to train the outlier score prediction layer. The remaining 500 points serve as the test set. The trained outlier score prediction layer is applied starting at time 1000, and prediction is performed by repeating one-step-ahead forecasting. Because future observations are not accessible during this phase, the predicted score at each step is used as part of the input to predict the next time point.

\subsubsection*{Beijing Temperature Data:}

The second dataset is taken from the PM2.5 Data of Five Chinese Cities~\cite{liang2016pm2}\footnote{\url{https://www.kaggle.com/datasets/uciml/pm25-data-for-five-chinese-cities}}. We use the temperature time series from Beijing for this experiment.

This dataset is divided into three parts. The first 9500 time points, which span approximately 13 months, are used to train the outlier detection layer. The following 1000 points are used for training the outlier score prediction layer, and the last 1000 points are reserved for testing.

\begin{figure}[t]
    \centering
    \includegraphics[width=1\linewidth]{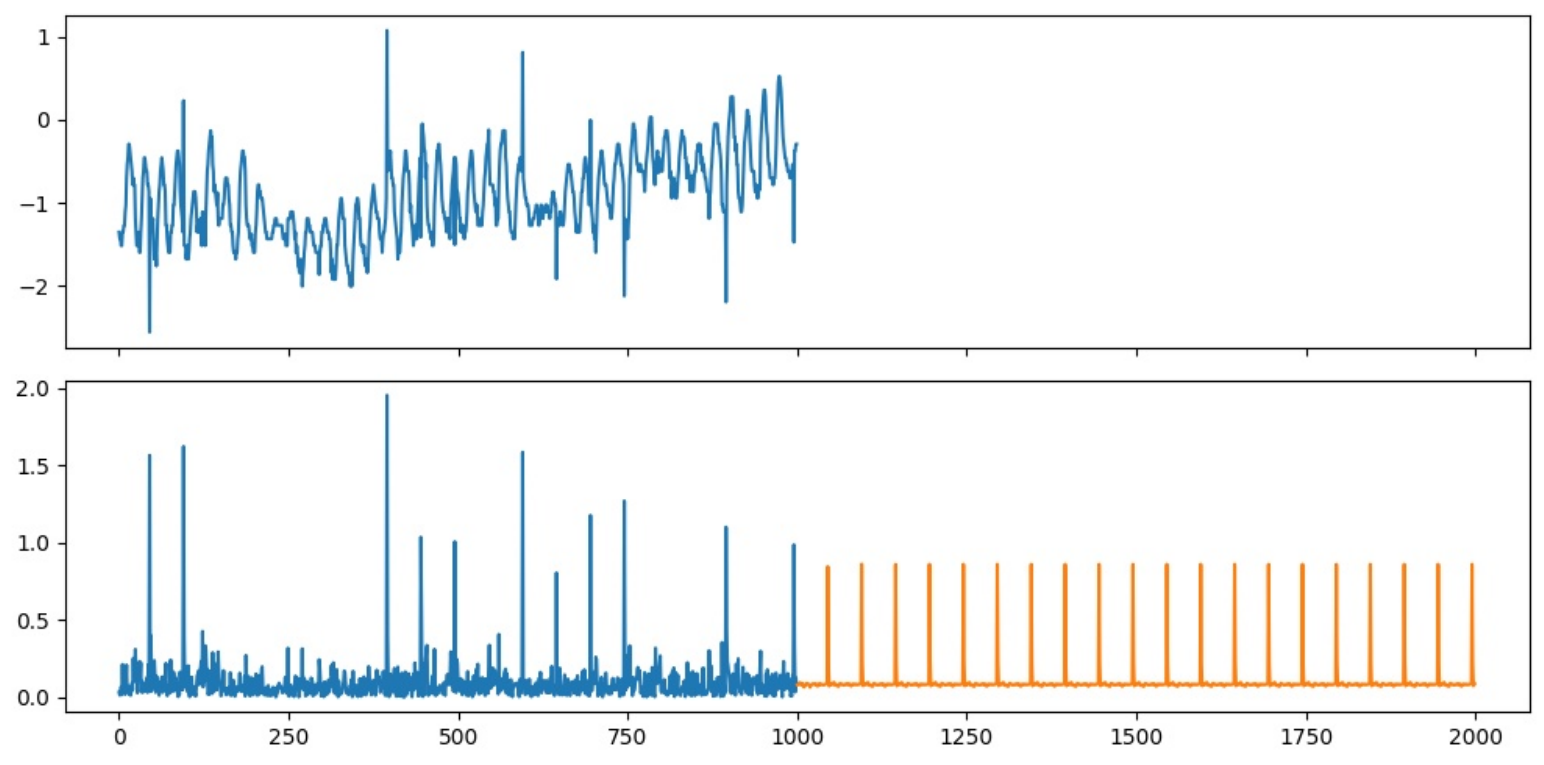}
    \caption{The Beijing temperature dataset used in our experiments. The top plot shows the original time series with artificially inserted periodic outliers. The bottom plot shows the outlier scores calculated by the detection layer (left half) and the predicted scores produced by the prediction layer (right half), starting at time point 1000.}
    \label{fig:beijing}
\end{figure}

To simulate periodic outliers, Gaussian noise is injected every 50 time steps. The noise is drawn from $\mathcal{N}(\text{std}/2, \text{std})$ with probability $1/2$, and from $\mathcal{N}(-\text{std}/2, \text{std})$ with probability $1/2$, where $\text{std}$ denotes the standard deviation of the entire dataset. Figure~\ref{fig:beijing} shows the resulting time series, along with the predicted outlier scores generated by the model.

\subsubsection*{Model Configuration:}

Both the outlier detection and outlier score prediction layers are implemented using simple neural networks, each consisting of a single LSTM layer. For the synthetic dataset, the detection layer uses a window size of 30 and outputs 64-dimensional representations. The prediction layer uses a window size of 50 and outputs 128-dimensional vectors. For the Beijing dataset, the detection layer uses a window size of 24, while the prediction layer again uses a window size of 50. In both layers, the output dimensionality is set to 256.

All models are trained using the Adam optimizer~\cite{kingma2014adam}.

\subsubsection*{Evaluation and result:}

In both the synthetic and real-world datasets, the timing of outlier occurrences is known in advance. This allows us to quantitatively evaluate the prediction accuracy of the proposed method using the area under the ROC curve (AUC) for the predicted outlier scores. While practical deployment would involve reporting an outlier when the predicted score exceeds a fixed threshold, we adopt AUC in this study to provide a threshold-independent evaluation.

The results show an AUC of $1.00$ in both datasets. This indicates that the proposed method can perfectly predict the occurrence of outliers over future time points. The periodicity of the inserted anomalies is 50 time steps, so accurate prediction requires the model to capture patterns that span at least that range. To verify this, we reduced the window size of the outlier score prediction layer to 40. As expected, the prediction accuracy dropped significantly, yielding an AUC of 0.44, which is close to random. In contrast, increasing the window size to 100 did not affect performance; the AUC remained at 1.00.

\begin{figure}[t]
    \centering
    \includegraphics[width=1\linewidth]{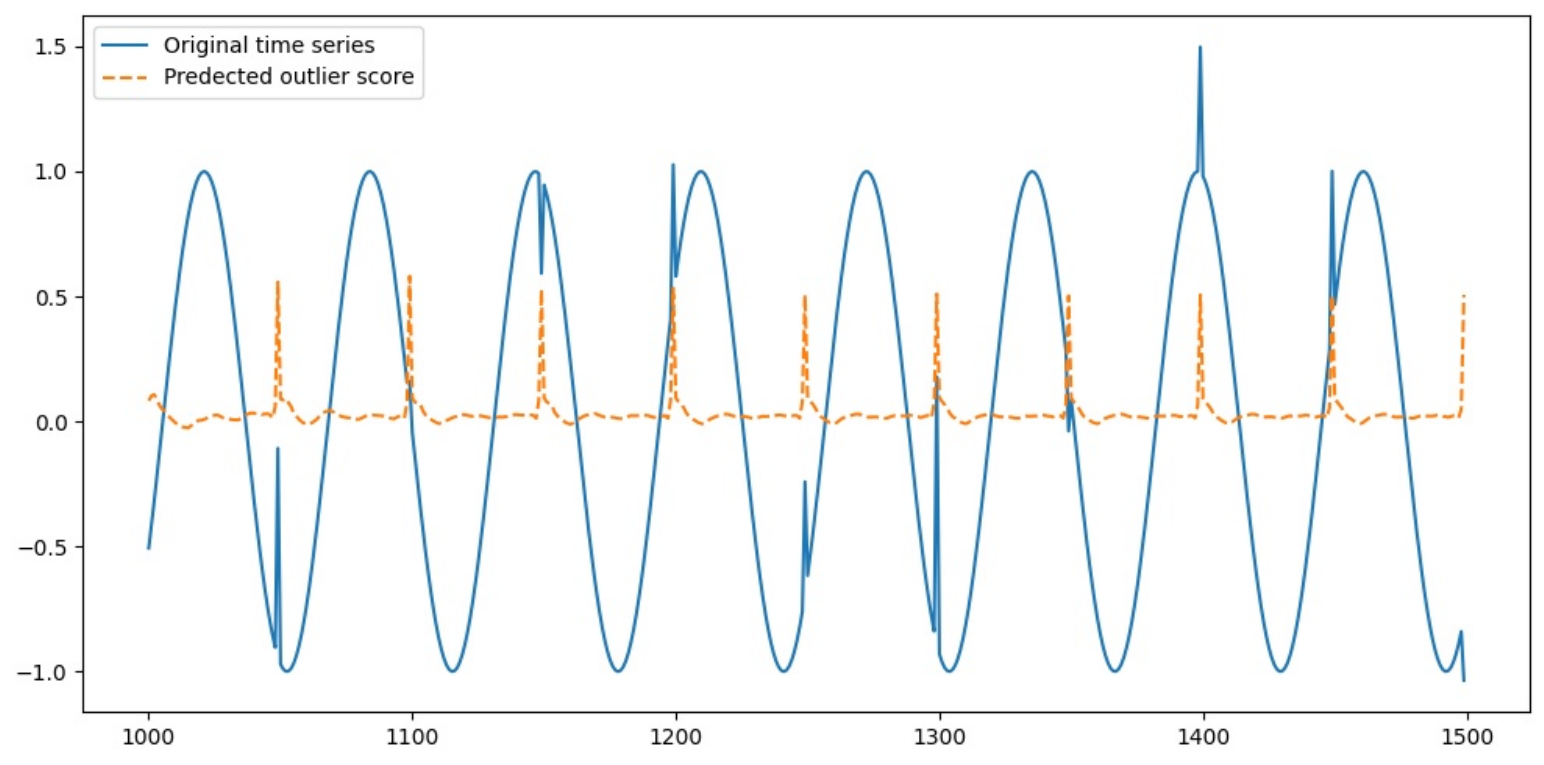}
    \caption{Predicted outlier scores for the synthetic time series in the test interval ($t = 1000$ to $1500$). The solid line shows the original time series with periodic anomalies. The dashed line shows the predicted outlier scores. Their peaks align with the injected outliers, indicating perfect long-term prediction.}
    \label{fig:exp1}
\end{figure}

\begin{figure}[t]
    \centering
    \includegraphics[width=1\linewidth]{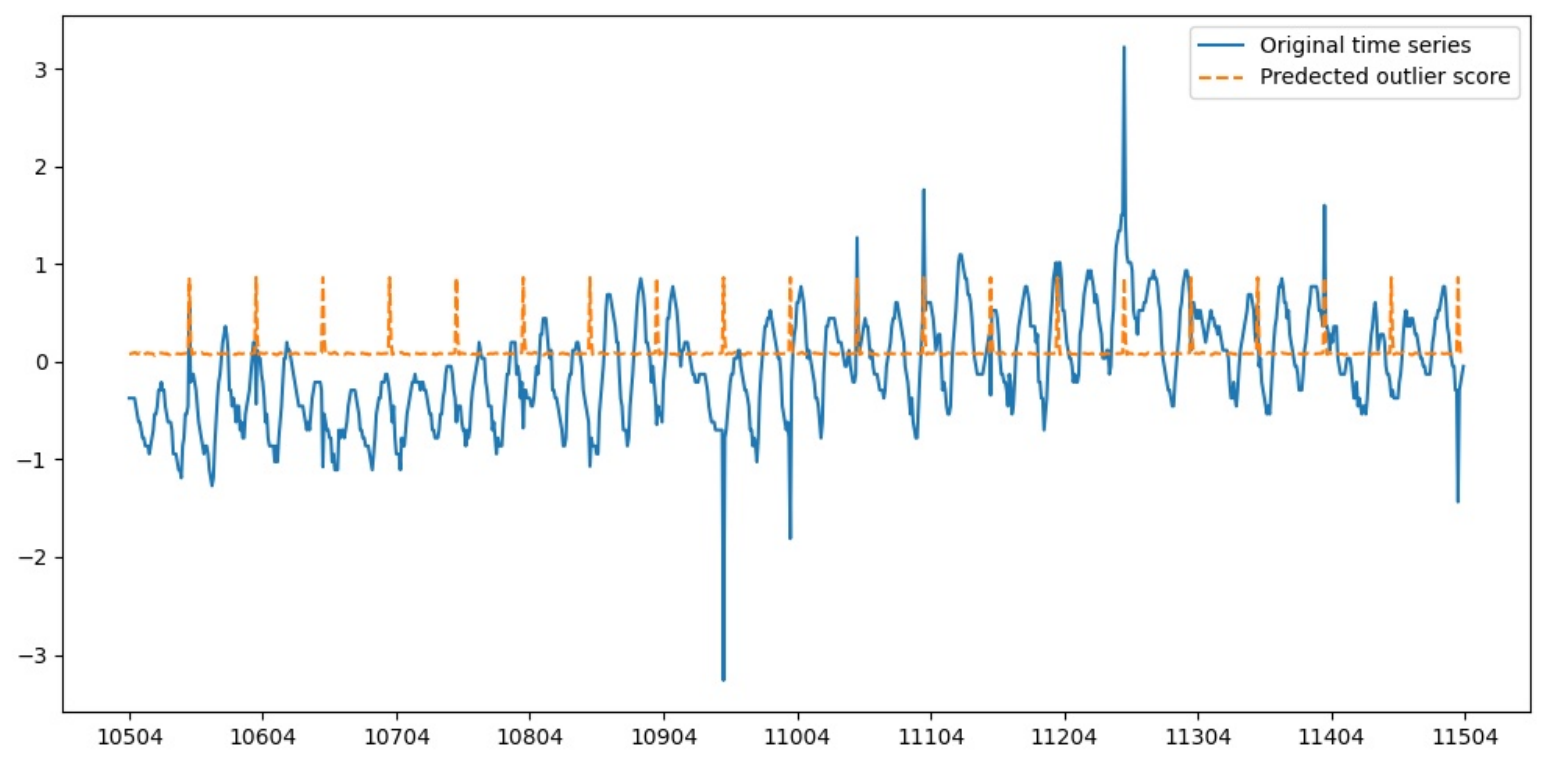}
    \caption{Predicted outlier scores for the Beijing temperature data during the test interval. The solid line represents the original series with periodic outliers, and the dashed line represents the predicted scores. The synchronization between the two indicates successful long-term forecasting. All values are normalized.}
    \label{fig:beijing2}
\end{figure}

Figure~\ref{fig:exp1} shows the original synthetic time series, which includes periodically inserted outliers, along with the predicted outlier scores for the test interval between time 1000 and 1500. The rise in predicted scores is aligned with the injected anomalies, confirming that the method correctly identifies the periodic structure. A similar trend is observed in the real dataset, as illustrated in Figure~\ref{fig:beijing2}. The dashed line indicates the predicted outlier scores, which consistently rise in accordance with the timing of inserted noise.

It is important to note that, during the test phase, the actual time series values are not available. This reflects a realistic setting in which future observations are unknown. Therefore, conventional outlier detection methods cannot produce such long-term forecasts, since they rely on the presence of observed data. In contrast, our method successfully extrapolates future outlier patterns based on learned temporal structures in the score sequences.

\subsection{Multivariate Time Series}

We now turn to a more complex scenario involving multivariate time series. In this setting, an outlier occurring in one variable may trigger a subsequent outlier in another variable. Our goal is to examine whether the proposed method can accurately predict such temporally and cross-dimensionally dependent outliers. To this end, we conduct experiments on both synthetic and real datasets, where correlated outliers are artificially inserted.

\subsubsection*{Experimental Setting}

We first consider a synthetic bivariate time series defined as follows:
\begin{align}
   x_{t}^{(1)} &= \sin(t/10), \\
   x_{t}^{(2)} &= \cos(t/10).
\end{align}
The sequence contains 1500 time points. The first 1000 points are used to train the proposed model, which consists of an outlier detection layer followed by an outlier score prediction layer. The remaining 500 points are reserved for testing.

Outliers are introduced according to the following procedure. We randomly select 30 time points from the entire sequence using a uniform distribution. At each selected time point $\tau$, Gaussian noise drawn from $\mathcal{N}(0, 0.5)$ is added to the second variable $x^{(2)}_{\tau}$. Then, 10 time steps later, a noise sampled from the same distribution is added to the first variable $x^{(1)}_{\tau+10}$. This setup creates a dependency in which an outlier in the second variable induces a delayed outlier in the first variable.

We also apply the same experimental framework to real-world data from the PM2.5 Data of Five Chinese Cities. We use the temperature and pressure time series from Beijing. Outliers are inserted into the temperature series at random time points, corresponding to approximately 1\% of the total data. Whenever an outlier is injected into the temperature series, a corresponding outlier is inserted into the pressure series exactly 10 steps later. Each injected outlier is implemented as a Gaussian perturbation. With equal probability, it follows either $\mathcal{N}(\text{std}, \text{std})$ or $\mathcal{N}(-\text{std}, \text{std})$, where $\text{std}$ denotes the standard deviation of the full dataset.

\begin{figure}[t]
    \centering
    \includegraphics[width=0.8\linewidth]{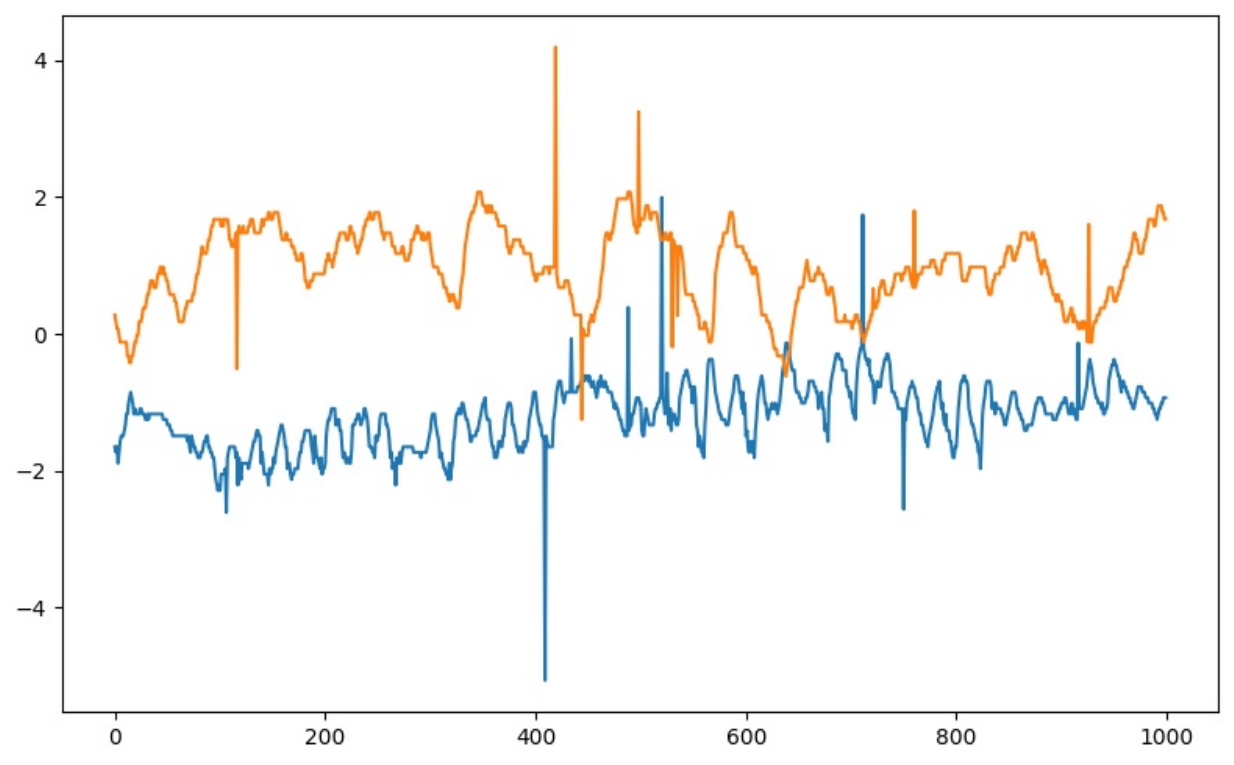}
    \caption{Bivariate real dataset (Beijing temperature and pressure) with correlated outliers. The blue and orange lines indicate temperature and pressure, respectively. The pressure series consistently shows outliers 10 time steps after the temperature spikes. All values are normalized.}
    \label{fig:realdata_multi_plot2}
\end{figure}

Figure~\ref{fig:realdata_multi_plot2} shows the temperature and pressure time series for the first 1000 points. The correlation between anomalies—specifically, that pressure spikes follow temperature anomalies by 10 time steps—is visually apparent.

As in the univariate experiments, we use neural networks with a single LSTM layer in both the outlier detection and prediction components. In the synthetic dataset, the output dimension of the LSTM is set to 64. For the real dataset, the outlier detection layer uses 512 dimensions, and the prediction layer uses 256 dimensions. Inputs and outputs in both cases are two-dimensional vectors corresponding to the bivariate nature of the data.

\subsubsection*{Evaluation and Results:}

As in the previous experiment, we evaluate the prediction accuracy using the area under the ROC curve (AUC). However, unlike the univariate case, the multivariate setting lacks periodicity in the timing of outlier occurrences. Therefore, evaluating the accuracy across the entire 500-point test sequence is not meaningful.

In this setting, outliers in the second variable occur randomly and without any warning signals. Consequently, they are fundamentally unpredictable. Outliers in the first variable are triggered by outliers in the second variable, but only occur once, exactly ten steps later. Since there is no further structure, it is not possible to predict any subsequent anomalies beyond that.

To evaluate prediction accuracy under these conditions, we use the following strategy. At each test time point $\tau$, we use the observed values and outlier scores up to $\tau$ to forecast the outlier score at time $\tau + k$, where $k$ is the prediction horizon. We repeat this procedure for a range of $k$ values and calculate the corresponding AUCs.

The expected behavior is as follows. For the second variable, which contains randomly occurring outliers, AUC should remain close to 0.5 regardless of $k$. For the first variable, prediction accuracy should be high when $k = 10$, corresponding to the known delay between cause and effect. When $k$ exceeds 10, performance is expected to decline rapidly.

Tables~\ref{table:bivariate1} and \ref{table:bivariate2} present the AUC values for various prediction horizons $k$ in the synthetic and Beijing datasets, respectively. In both cases, the first variable achieves perfect or near-perfect prediction up to $k = 10$, and becomes nearly unpredictable beyond that point. As anticipated, the second variable remains close to random prediction throughout.

\begin{figure}[t]
    \centering
    \includegraphics[width=1\linewidth]{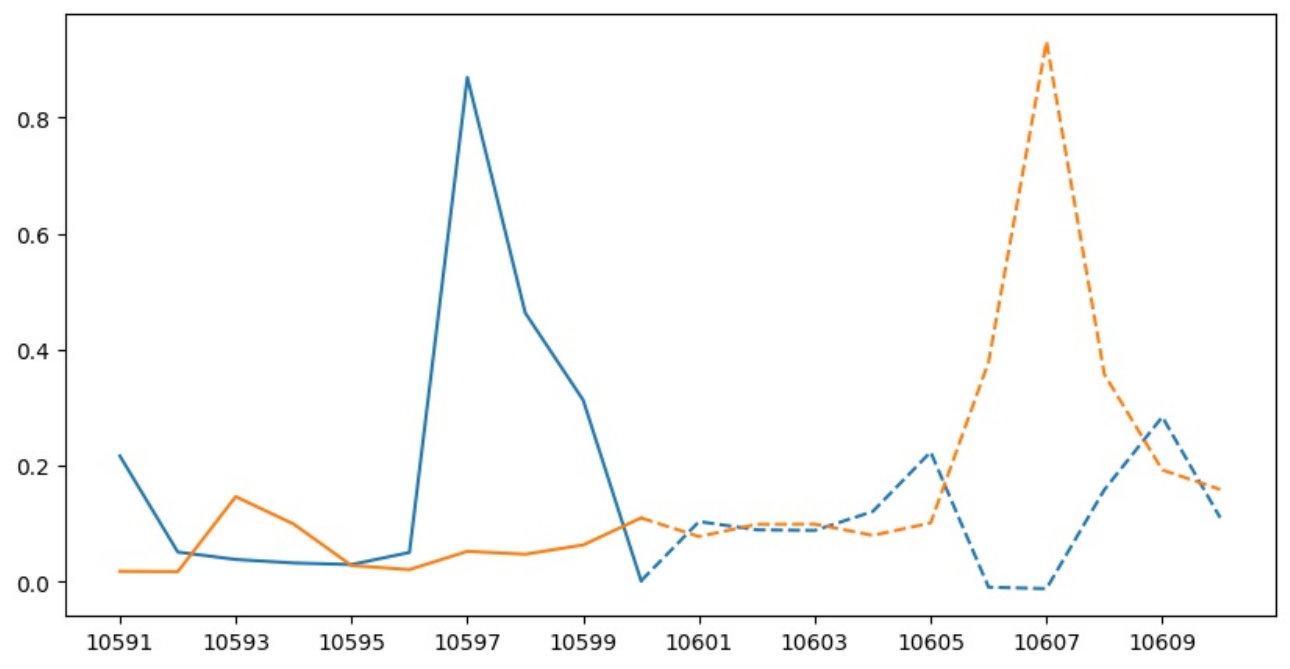}
    \caption{Example of long-term outlier score prediction in the Beijing bivariate dataset. The prediction is made at $T = 10600$. The solid lines show the true outlier scores before prediction time. A spike in the temperature score at $10597$ indicates an outlier. The dashed lines show the predicted scores, and the rise in the pressure score at $10607$ confirms successful forecasting.}
    \label{fig:exp2}
\end{figure}

Figure~\ref{fig:exp2} shows a concrete example using the Beijing dataset. At time $T = 10600$, we make a prediction of future outlier scores. The values before this point are based on actual observations, while those after are model predictions. At time $10597$, an anomaly is detected in the temperature series. Ten steps later, at time $10607$, an anomaly is predicted in the pressure series. The rise in the predicted outlier score for pressure at $10607$ is clearly visible and correctly aligned with the delayed effect.

Interestingly, in both datasets, AUC values for the second variable sometimes fall slightly below 0.5 when $k$ is small. This counterintuitive result may arise from temporal autocorrelation in the outlier scores themselves. After an outlier occurs, the score may remain elevated for a while, even when no further outlier follows. This behavior leads the model to produce false positives, lowering the AUC.

\begin{table}[t]
    \caption{Long-term prediction accuracy for the synthetic bivariate time series. The AUC for each prediction horizon $k$ is shown. The first variable is predictable up to $k = 10$ due to the known delay, while the second variable remains unpredictable.}
    \label{table:bivariate1}
    \centering
    \vspace{4mm}
    \begin{tabular}{|c|c|c|c|c|c|c|c|c|c|c|c|c|c|c|}
    \hline
    Prediction horizon $k$ & 1 & 2 & 3 & 4 & 5 & 6 & 7 & 8 & 9 & 10 & 11 & 12 & 13 & 14 \\ \hline \hline
    Variable $x^{(1)}$ & 1.00 & 1.00 & 1.00 & 1.00 & 1.00 & 1.00 & 1.00 & 1.00 & 1.00 & 1.00 & 0.45 & 0.45 & 0.49 & 0.49 \\ \hline
    Variable $x^{(2)}$ & 0.40 & 0.39 & 0.39 & 0.42 & 0.43 & 0.44 & 0.45 & 0.48 & 0.50 & 0.51 & 0.49 & 0.45 & 0.46 & 0.46 \\ \hline
    \end{tabular}
    \vspace{2mm}
\end{table}

\begin{table}[t]
    \caption{Long-term prediction accuracy for the Beijing bivariate time series. The AUC for each prediction horizon $k$ is shown. As in the synthetic data, the pressure series is predictable up to $k = 10$ due to its causal dependency, while the temperature series remains unpredictable.}
    \label{table:bivariate2}
    \centering
    \vspace{4mm}
    \begin{tabular}{|c|c|c|c|c|c|c|c|c|c|c|c|c|c|c|}
    \hline
    Prediction horizon $k$ & 1 & 2 & 3 & 4 & 5 & 6 & 7 & 8 & 9 & 10 & 11 & 12 & 13 & 14 \\ \hline \hline
    Pressure & 0.83 & 0.87 & 0.96 & 0.87 & 0.87 & 0.97 & 0.97 & 0.97 & 0.97 & 0.96 & 0.52 & 0.45 & 0.48 & 0.47 \\ \hline
    Temperature & 0.41 & 0.34 & 0.39 & 0.40 & 0.43 & 0.46 & 0.49 & 0.50 & 0.49 & 0.49 & 0.49 & 0.49 & 0.48 & 0.45 \\ \hline
    \end{tabular}
    \vspace{2mm}
\end{table}

\section{Conclusions and Future Work}

This study investigated the relatively unexplored problem of long-term outlier prediction in time series data. Outlier detection plays an important role in applications such as network security and industrial monitoring. However, most conventional methods focus on identifying immediate anomalies by detecting deviations from recent norms. These approaches are not designed to anticipate outlier occurrences far into the future, which limits their practical value in settings where long-term forecasting is essential. To bridge this gap, we proposed a new unsupervised framework for predicting future outliers based on past detection results.

The proposed method is based on a simple two-layer architecture. The first layer detects outliers at each time point by applying any existing outlier detection method to the observed time series. The second layer models the temporal structure of the detected outlier scores and predicts future scores accordingly. This framework does not rely on a specific model and can be combined with various detection and forecasting techniques.

We conducted experiments using both synthetic and real datasets, where correlated outliers were injected to simulate realistic scenarios. The results showed that our approach can effectively capture patterns in outlier occurrence and achieve accurate long-term predictions. These findings suggest that the method provides a useful baseline for future research in this area.

Several directions remain open for future work. One important extension is to explore alternative modeling strategies for both the detection and prediction layers. Since outlier scores often exhibit spiky, event-driven patterns, point process models may be more appropriate than standard time series predictors. Another promising direction is to develop online learning methods for long-term outlier forecasting. Such models would be continuously updated using streaming data and adapt to changing environments. Finally, analyzing the causal relationships among outliers and improving the interpretability of the predictions are also important challenges. These directions would enhance both the practicality and the explanatory power of long-term outlier prediction methods.

\bibliographystyle{splncs04}
\bibliography{references}

@article{hodge2004survey,
  title={A survey of outlier detection methodologies},
  author={Hodge, Victoria and Austin, Jim},
  journal={Artificial Intelligence Review},
  volume={22},
  pages={85--126},
  year={2004}
}

@article{boukerche2020outlier,
  title={Outlier detection: Methods, models, and classification},
  author={Boukerche, Azzedine and Zheng, Lining and Alfandi, Omar},
  journal={ACM Computing Surveys (CSUR)},
  volume={53},
  number={3},
  pages={1--37},
  year={2020},
}

@article{blazquez2021review,
  title={A review on outlier/anomaly detection in time series data},
  author={Bl{\'a}zquez-Garc{\'\i}a, Ane and Conde, Angel and Mori, Usue and Lozano, Jose A},
  journal={ACM computing surveys (CSUR)},
  volume={54},
  number={3},
  pages={1--33},
  year={2021}
}

@journal{intro1,
  author    = {Manish Gupta and Jing Gao and Charu C. Aggarwal and Jiawei Han},
  title     = {Outlier Detection for Temporal Data: A Survey},
  journal = {IEEE Transactions on Knowledge and Data Engineering (TKDE)},
  pages     = {2250--2267},
  year      = {2014}
}

@article{intro2,
  author    = {Mohammad Braei and Sebastian Wagner},
  title     = {ANOMALY DETECTION IN UNIVARIATE TIME-SERIES: A SURVEY ON THE STATE-OF-THE-ART},
  journal = {arXiv},
  year      = {2020}
}

@inproceedings{breunig2000lof,
  title={{LOF}: Identifying Density-based Local Outliers},
  author={Breunig, Markus M and Kriegel, Hans-Peter and Ng, Raymond T and Sander, J{\"o}rg},
  booktitle={Proceedings of the ACM SIGMOD International Conference on Management of Data (SIGMOD)},
  pages={93--104},
  year={2000}
}

@inproceedings{yamanishi2002unifying,
  title={A Unifying Framework for Detecting Outliers and Change Points from Non-stationary Time Series Data},
  author={Yamanishi, Kenji and Takeuchi, {Jun-ichi}},
  booktitle={Proceedings of the Eighth ACM SIGKDD International Conference on Knowledge Discovery and Data Mining (KDD)},
  pages={676--681},
  year={2002}
}

@inproceedings{sakurada2014anomaly,
  title={Anomaly Detection Using Autoencoders with Nonlinear Dimensionality Reduction},
  author={Sakurada, Mayu and Yairi, Takehisa},
  booktitle={Proceedings of the 2014 Second Workshop on Machine Learning for Sensory Data Analysis (MLSDA)},
  pages={4--11},
  year={2014}
}

@article{chandola2009anomaly,
  title={Anomaly detection: A Survey},
  author={Chandola, Varun and Banerjee, Arindam and Kumar, Vipin},
  journal={ACM Computing Surveys (CSUR)},
  volume={41},
  number={3},
  pages={1--58},
  year={2009}
}

@article{chalapathy2019deep,
  title={Deep Learning for Anomaly Detection: A Survey},
  author={Pang, Guansong and Shen, Chunhua and Cao, Longbing and Hengel, Anton Van Den},
  journal={ACM Computing Surveys (CSUR)},
  volume={54},
  number={2},
  pages={1--38},
  year={2021}
}

@book{hamilton2020time,
  title={Time Series Analysis},
  author={Hamilton, James D},
  year={2020},
  publisher={Princeton University Press}
}

@book{shumway2000time,
  title={Time Series Analysis and its applications},
  author={Shumway, Robert H and Stoffer, David S and Stoffer, David S},
  volume={3},
  year={2000},
  publisher={Springer}
}

@article{hochreiter1997long,
  title={Long Short-term Memory},
  author={Hochreiter, Sepp and Schmidhuber, J{\"u}rgen},
  journal={Neural Computation},
  volume={9},
  number={8},
  pages={1735--1780},
  year={1997}
}

@inproceedings{cho2014learning,
  title={Learning Phrase Representations Using RNN Encoder--Decoder for Statistical Machine Translation},
  author={Cho, Kyunghyun and van Merrienboer, Bart and Gulcehre, Caglar and Bahdanau, Dzmitry and Bougares, Fethi and Schwenk, Holger and Bengio, Yoshua},
  booktitle={Proceedings of the 2014 Conference on Empirical Methods in Natural Language Processing (EMNLP)},
  pages={1724},
  year={2014}
}

@article{wen2022transformers,
  title={Transformers in Time Series: A Survey},
  author={Wen, Qingsong and Zhou, Tian and Zhang, Chaoli and Chen, Weiqi and Ma, Ziqing and Yan, Junchi and Sun, Liang},
  journal={arXiv preprint arXiv:2202.07125},
  year={2022}
}

@article{lim2021time,
  title={Time-series Forecasting with Deep Learning: A Survey},
  author={Lim, Bryan and Zohren, Stefan},
  journal={Philosophical Transactions of the Royal Society A},
  volume={379},
  number={2194},
  year={2021}
}

@article{kingma2014adam,
  title={Adam: A method for stochastic optimization},
  author={Kingma, Diederik P and Ba, Jimmy},
  journal={arXiv preprint arXiv:1412.6980},
  year={2014}
}

@article{liang2016pm2,
  title={{PM}2. 5 data reliability, consistency, and air quality assessment in five Chinese cities},
  author={Liang, Xuan and Li, Shuo and Zhang, Shuyi and Huang, Hui and Chen, Song Xi},
  journal={Journal of Geophysical Research: Atmospheres},
  volume={121},
  number={17},
  pages={10--220},
  year={2016}
}
\end{document}